\documentclass[sigconf]{acmart}

\usepackage{amsmath,amsfonts}
\usepackage{booktabs}
\usepackage{hyperref}
\usepackage{xcolor}
\usepackage{color}
\usepackage[dvipsnames]{xcolor}

\begin{document}

\title[The (R)evolution of Scientific Workflows in the Agentic AI Era: Towards Autonomous Science]{The (R)evolution of Scientific Workflows in the Agentic AI Era: \\ Towards Autonomous Science}

\settopmatter{authorsperrow=4} 
\author[W. Shin]{Woong Shin}
\affiliation{
  \institution{Oak Ridge National Lab.}
  \city{Oak Ridge, TN}
  \country{USA}
}
\author[R. Souza]{Renan Souza}
\affiliation{
  \institution{Oak Ridge National Lab.}
  \city{Oak Ridge, TN}
  \country{USA}
}
\author[D. Rosendo]{Daniel Rosendo}
\affiliation{
  \institution{Oak Ridge National Lab.}
  \city{Oak Ridge, TN}
  \country{USA}
}
\author[F. Suter]{Frédéric Suter}
\affiliation{
  \institution{Oak Ridge National Lab.}
  \city{Oak Ridge, TN}
  \country{USA}
}
\author[F. Wang]{Feiyi Wang}
\affiliation{
  \institution{Oak Ridge National Lab.}
  \city{Oak Ridge, TN}
  \country{USA}
}
\author[P. Balaprakash]{Prasanna Balaprakash}
\affiliation{
  \institution{Oak Ridge National Lab.}
  \city{Oak Ridge, TN}
  \country{USA}
}
\author[R. Ferreira da Silva]{Rafael Ferreira da Silva}
\affiliation{
  \institution{Oak Ridge National Lab.}
  \city{Oak Ridge, TN}
  \country{USA}
}

\thanks{Notice: This manuscript has been authored in part by UT-Battelle, LLC under Contract No. DE-AC05-00OR22725 with the U.S. Department of Energy. The United States Government retains and the publisher, by accepting the article for publication, acknowledges that the United States Government retains a non-exclusive, paid-up, irrevocable, world-wide license to publish or reproduce the published form of this manuscript, or allow others to do so, for United States Government purposes. The Department of Energy will provide public access to these results of federally sponsored research in accordance with the DOE Public Access Plan (http://energy.gov/downloads/doe-public-access-plan).}

\begin{abstract}
Modern scientific discovery increasingly requires coordinating distributed facilities and heterogeneous resources, forcing researchers to act as manual workflow coordinators rather than scientists. Advances in AI leading to AI agents show exciting new opportunities that can accelerate scientific discovery by providing intelligence as a component in the ecosystem. However, it is unclear how this new capability would materialize and integrate in the real world. To address this, we propose a conceptual framework where workflows evolve along two dimensions which are intelligence (from static to intelligent) and composition (from single to swarm) to chart an evolutionary path from current workflow management systems to fully autonomous, distributed scientific laboratories. With these trajectories in mind, we present an architectural blueprint that can help the community take the next steps towards harnessing the opportunities in autonomous science with the potential for 100x discovery acceleration and transformational scientific workflows.
\end{abstract}


\maketitle

\section{Introduction}

Modern scientific discovery is undergoing a profound shift as the scale, speed, and complexity of research increase. Addressing urgent challenges in areas like materials design, climate modeling, and health science now demands seamless coordination across many geographically distributed and technologically diverse facilities~\cite{antypas2021enabling}. For instance, a typical materials discovery campaign may span over ten facilities, including synthesis labs, user facilities, and high-performance computing (HPC) centers, and require months of manual coordination~\cite{abolhasani2023rise}. This operational overhead limits the pace of progress and forces researchers to act less as scientists and more as orchestrators of workflows. The long-standing vision of fully autonomous science offers a way forward: systems where instruments, robots, computational models, and data pipelines operate continuously and intelligently. By embedding reasoning and adaptation into workflows, these labs have the potential to accelerate discovery by factors of 10 to 100, transforming exploratory science into a continuous, machine-augmented process~\cite{autonomousscience}.

Artificial intelligence (AI) has already revolutionized industrial automation, enabling fully autonomous systems in manufacturing, supply chain logistics, and even autonomous vehicles~\cite{ribeiro2021robotic}. These advances open up exciting new possibilities for advancing scientific research. Recent efforts such as self-driving chemistry labs, foundation model integration, and near real-time simulation steering suggest that AI agents can serve as central coordinators across experimental and computational platforms~\cite{abolhasani2023rise}. However, scientific workflows pose challenges that go beyond those found in commercial automation. Scientific decision-making must remain transparent, reproducible, and grounded in physical principles. Workflows must integrate complex instruments and data modalities, and support collaboration among domain scientists, facility operators, and computational teams~\cite{ferreiradasilva2024computer}. Commercial solutions, while powerful, are often brittle when applied to the unique constraints of scientific discovery. As a result, new approaches are needed that balance automation with trust, adaptability with control, and scalability with human oversight.

Existing workflow management systems (WMSs) have delivered robust execution frameworks that enable reproducible data pipelines across HPC and cloud platforms~\cite{suter2025fgcs}. Yet these systems are not designed to reason about scientific goals, adapt to new data in near real-time, or coordinate across physically distributed agents and instruments. AI integration into workflows remains mostly ad hoc, with isolated prototypes and custom interfaces that risk fragmenting the ecosystem. Without a systematic approach to combining workflow systems with intelligent agent infrastructure, we risk stagnation and the erosion of years of hard-won community infrastructure. To realize the vision of autonomous science~\cite{autonomousscience}, we must evolve these tools into platforms that support intelligent, multi-agent orchestration, enable provenance-aware decision-making, and offer robust coordination across laboratory and computing environments~\cite{pauloski2025empowering}. The shift towards autonomous discovery should be an evolution rather than a revolution.

In this paper, we focus on identifying an evolutionary path towards enabling the vision of autonomous science building on top of the 20+ years of effort of the workflows community. To achieve this, we first aim to understand the fundamental relationship between existing workflows and emerging agentic AI systems. We identify that both traditional workflows and modern AI agents share the state machine abstraction as a common foundation for autonomy and use this to identify an evolutionary path in between. We propose that workflows can evolve along two dimensions: \textbf{intelligence} (static to intelligent) and \textbf{composition} (single to swarm) forming an evolutionary plane that enables workflow designers, system designers, and policymakers to reason about concrete step differences from traditional workflows to AI-driven autonomous discovery.

To this end, our contribution is threefold:
\begin{itemize}
    \item \textbf{Conceptual Framework for Evolution:} We present a conceptual framework that unifies traditional workflows and AI agents, revealing evolutionary paths that facilitate concrete roadmaps from traditional workflows to autonomous science.
    
    \item \textbf{Architectural Blueprint:} We provide an architectural blueprint that envisions autonomous scientific laboratories that materialize this evolution and demonstrate how scientific discovery would evolve with more autonomy.
    
    \item \textbf{Roadmap:} With the projected future opportunities that will be enabled by fully autonomous science, we identify concrete challenges and the strategic bets required in both the AI community and the scientific workflows community.
\end{itemize}

The remainder of this paper is organized as follows. Section 2 provides background on scientific workflow complexity and reviews AI advances. Section 3 presents our evolution framework with its two key dimensions. Section 4 presents an architectural vision for systems embodying this evolution. Section 5 identifies challenges and opportunities in realizing autonomous scientific libraries. Section 6 discusses implications, limitations, and future work. Section 7 concludes with our vision and call to action for the community.

\section{Background and Motivation}
\label{sec:background}

\subsection{Scientific Workflow Systems \\ and Their Limits}

Scientific workflows can be formally defined as structured compositions of computational tasks to achieve a research objective~\cite{suter2025fgcs}. The most widespread structure is that of a directed acyclic graph (DAG) whose nodes are the computational tasks and edges express data and control dependencies between them. Over the past decades, workflows have become the predominant format for describing complex, multi-step, multi-domain scientific applications. To manage their composition, the planning and orchestration of their execution on distributed computing infrastructures, a large number of WMSs have been proposed~\cite{workflow-systems}. However, modern workflows now include conditional branches, cycles, and human-in-the-loop components that require rethinking their design and execution beyond traditional and limited DAG structures~\cite{ferreiradasilva2024computer}. Moreover, scientific workflows are evolving from their traditional task-driven nature to embrace data-driven analytical and AI pipelines~\cite{suter2023escience}.

Modern workflow management systems have demonstrated strong capabilities in orchestrating large-scale data movement, ensuring fault tolerance, and managing resource provisioning across distributed environments. Established systems~\cite{suter2025fgcs} provide mature tooling for composing and executing complex scientific pipelines. These systems offer robust mechanisms for tracking task dependencies, handling failures, and optimizing performance across HPC and cloud platforms. Despite their reliability and maturity, they operate under a foundational constraint: workflows are typically represented as static DAGs that must be fully defined before execution. This limits their ability to respond to emerging data, evolving hypotheses, or near real-time system conditions. In the context of our evolutionary framework, these capabilities primarily fall within the Static and Adaptive regions, where the transition logic remains largely predetermined and fixed at design time. While sufficient for many current scientific workflows, these systems do not inherently support learning, optimization, or reasoning required for autonomous scientific operation.

\vspace{-\baselineskip}
\subsection{Multi-Facility Coordination Challenges}
The need to coordinate across multiple, heterogeneous facilities introduces additional complexity that static or even adaptive workflows struggle to address. Contemporary scientific campaigns increasingly rely on orchestrating activities across a continuum of resources, including HPC centers, experimental instruments, edge services, and storage platforms. Today's prevailing approach involves bespoke, facility-specific workflows stitched together through manual coordination~\cite{antypas2021enabling}. This ad hoc strategy is both brittle and labor-intensive, often requiring researchers to oversee synchronization and data handoffs between systems. A representative example is a materials discovery campaign that cycles between synthesis at a user facility, characterization at a beamline, and simulation on an HPC system~\cite{abolhasani2023rise}. As the number of facilities, stakeholders, and interdependencies increases, the coordination overhead grows rapidly, consuming valuable time and human effort. These challenges highlight the limitations of current workflow abstractions and underscore the need for systems that can reason about and adapt to multi-facility execution contexts dynamically and autonomously.

A growing challenge in modern scientific workflows is the orchestration of multi-facility deployments that span experimental instruments, edge computing resources, and centralized HPC systems. These workflows increasingly operate across the Edge–Cloud–HPC continuum, reflecting the complex, distributed nature of contemporary scientific discovery campaigns~\cite{tyler2022cross, skluzacek2024towards}. While current workflow systems are effective within single-institution contexts, they lack native mechanisms for coordinating across heterogeneous environments with differing interfaces, policies, and operational constraints. This often results in error-prone, manually maintained integration layers that introduce significant overhead and reduce scientific agility. As workflows evolve to support more adaptive and intelligent behaviors, enabling seamless, near real-time coordination across facilities will be essential for realizing autonomous science at scale.

\subsection{AI Advances and Opportunities}

In the industry, the ChatGPT phenomenon ignited unprecedented excitement about the transformative potential of AI, with 400 million weekly active users by 2025 and adoption by 80\% of Fortune 500 companies~\cite{chatgpt2025, openai2023enterprise}. This ignition represents a new AI boom beyond traditional machine learning (ML), as emerging autonomous agents that can handle complex tasks independently~\cite{ibm2025ai}. By 2025, 99\% of enterprise developers are exploring AI agents with platforms like AWS AgentCore and Microsoft Azure enabling autonomous task execution~\cite{ibm2025ai}. Demonstrating AI evolution from assistive to autonomous systems, these efforts leverage large language models (LLMs) for reasoning, neural networks for pattern recognition, and multi-agent coordination for complex tasks.

In science, advances in AI are revealing new opportunities in accelerating scientific discovery. Early efforts integrated ML for parameter prediction and hyperparameter tuning~\cite{8638041,wozniak2018candle}. Later, autonomous experimentation platforms have emerged. Berkeley A-lab processes 50-100 times more samples than humans daily, synthesizing 41 novel materials in 17 days~\cite{szymanski2023autonomous}, while self-driving laboratories like Ada optimize thin films~\cite{macleod2020self} and ChemOS orchestrates distributed~\cite{roch2020chemos}. Recent approaches leverage LLMs. For example, ChemChow integrates GPT-4 with 18 chemistry tools to autonomously plan synthesis~\cite{bran2024augmenting}. Despite progress, these efforts remain point solutions and lack systematic integration. From a scientific workflow perspective, moving from current systems to such autonomous solutions is a disruptive disjoint leap.

Scientific workflows operate under fundamentally different constraints from the industry, prioritizing validation and reproducibility over pure efficiency~\cite{abolhasani2023rise}. On top of this, the physical nature of experimental sciences imposes physical constraints such as irreplaceable samples, expensive equipment, and irreversible experiments. Further, multi-stakeholder complexity in a multi-facility scientific environment compounds the challenges. Scientific discovery demands understanding causality beyond pattern recognition, integrating theory with experiment, and maintaining detailed provenance for reproducibility~\cite{deelman2023artificial}. These requirements make it necessary to have a domain-specific approach to AI integration that bridges proven workflow infrastructure with emerging autonomous capabilities.

\subsection{The Integration Challenge}

Despite the exciting advancements in AI, there is a gap between traditional workflow capabilities and materializing the AI potential. The scientific workflow community faces a critical integration challenge~\cite{badia2024integrating}. Without a systematic approach to bridge these paradigms, the community risks either stagnation or chaotic disruption, missing the opportunity to evolve towards a coherent vision of autonomous scientific laboratories. Traditional WMSs and emerging AI technologies exist in separate worlds, with fragmented ad-hoc one-off integration attempts far from being reproducible~\cite{nouri2021exploring, brewer2025, deelman2023artificial}. Further, it is unclear how the non-deterministic nature demonstrated by the advent of GenAI would even be useful in scientific endeavors that require high levels of determinism. Due to the ongoing challenges even without AI, neither abandoning proven workflow infrastructure nor ignoring the transformative AI potential is viable. Yet, it is difficult to reason about how much of the infrastructure would be able to adopt AI capabilities and how much additional investment one would need.

To address these issues and pave an evolutionary trajectory, we need a conceptual framework that helps us reason about traditional and AI-based approaches in a unified way. This framework needs to reveal practical migration paths that maintain scientific requirements such as reproducibility, validation, and provenance, while gradually adopting new capabilities. It should guide both technical development and community adoption by revealing incremental but transformative transitions. It should naturally bridge the transition from current WMSs to the ultimate goal of fully autonomous scientific discovery by providing a roadmap that preserves investments while enabling new capabilities.

\section{The Evolution Framework}
\label{sec:framework}

In this section, we present a systematic approach to understanding how scientific workflows can consider a progression towards AI in a continuum of evolution instead of a revolution. Our key insight is that both workflows and AI systems can be reduced down to an autonomous primitive, an agent, which is \textit{anything that can be viewed as perceiving its environment through sensors and acting upon that environment through actuators}~\cite{russell2021artificial}. This primitive can be modeled as state machines as a common denominator, and with varying levels of sophistication in their transition functions and composition patterns. This forms a two-dimensional spectrum of autonomy from static workflows to multi-agent coordination that captures the complexity of fully autonomous systems as a natural progression in intelligence and their composition.

\begin{figure}[ht!]
  \centering
  \includegraphics[width=7.5cm]{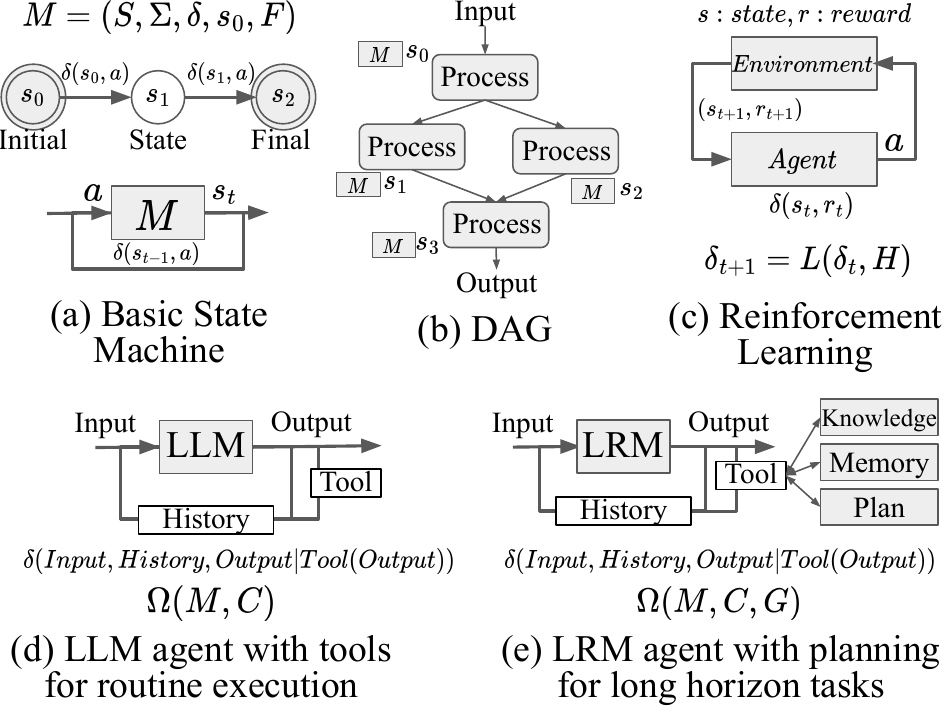}
  \caption{State machine abstraction as a common denominator across
  various types of autonomy: the extension of $\delta$ with learning $L$, optimization $\arg\min J$, or meta opt. $\Omega$  defines its sophistication}
  \label{figs:state_machine}
  \vspace{-4.0mm}
\end{figure}

\subsection{Autonomy: State Machine Abstraction}

The execution model of scientific workflows can be expressed as finite state machines $M = (S, \Sigma, \delta, s_0, F)$ (Figure~\ref{figs:state_machine}-a), where $S$ represents workflow stages, $\Sigma$ denotes the input alphabet of events and data, $\delta: S \times \Sigma \rightarrow S$ defines deterministic transitions, $s_0 \in S$ is the initial state, and $F \subseteq S$ represents final states. Figure~\ref{figs:state_machine}-b demonstrates how a DAG workflow directly maps to this model. Nodes correspond to states that represent computational tasks or data transformations, and edges encode the transition function based on task completion events. Instead of focusing on the DAG, this representation focusing on the execution unit of workflows, the state machine loop, provides a foundation for reasoning about workflow behavior, composition, and their extension. The deterministic nature of $\delta$ in traditional workflows ensures reproducibility but limits adaptability, which may no longer be suitable for coordinating distributed facilities and heterogeneous resources~\cite{antypas2021enabling, abolhasani2023rise}.

Modern AI agents, despite their apparent complexity, operate on the same state machine principles but with enhanced transition functions. The key distinction is in extending the basic formulation with mechanisms that provide dynamic behavior. Adaptive systems use $\delta: S \times \Sigma \times O \rightarrow S$ where $O$ represents observations or feedback signals. Learning systems employ $\delta_{t+1} = L(\delta_t, H)$ where $L$ is the learning function and $H$ is history, and intelligent systems implement $M' = \Omega(M, C, G)$ where $\Omega$ is a meta-optimization operator that can redefine the entire state machine ($M'$) based on context $C$ and mutable goals $G$. Figure~\ref{figs:state_machine}-c shows how a traditional machine learning (ML)-based system can be modeled in this way. Figure~\ref{figs:state_machine}-d models a large language model (LLM)- or large reasoning model (LRM)-based AI agent with tools that implement routine sequence tasks with some adaptability. Figure~\ref{figs:state_machine}-e models a more advanced LLM- or LRM-based AI agent that can learn, reason, plan, and execute tasks given the evolving environment while pursuing optimality. These dynamic entities at the lowest level then can be composed to implement higher-order complex emergent behaviors. Such a dynamic nature of AI-driven autonomy is the opportunity for the future of scientific discovery.

\subsection{The Intelligence Dimension}
\label{subsec:intelligence_dim}

To capture the evolution of dynamic behavior necessary, we define the intelligence dimension as the progressive sophistication of the transition function, establishing five evolutionary levels (Table~\ref{tab:intelligence}). 

\begin{table}[ht!]
  \caption{The intelligence dimension}
  \label{tab:intelligence}
  \footnotesize
  \begin{tabular}{p{3.8cm}p{4.0cm}}
    \toprule
    \textbf{Dimension} & \textbf{Description} \\
    \midrule
    \textbf{Static:} $\delta: S \times \Sigma \rightarrow S$ & 
    Transition function depends solely on current state and input, implementing predetermined execution paths \\
    \midrule
    \textbf{Adaptive:} $\delta: S \times \Sigma \times O \rightarrow S$ & 
    Extended with observations/feedback signals $O$ enabling runtime adjustments and conditional branching \\
    \midrule
    \textbf{Learning:} $\delta_{t+1} = L(\delta_t, H)$ & 
    Incorporates history through learning function $L$ that updates transitions based on experience $H$ \\
    \midrule
    \textbf{Optimizing:} $\delta^* = \arg\min_\delta J(\delta)$ & 
    Seeks optimal behavior via cost function $J$, balancing exploration and exploitation \\
    \midrule
    \textbf{Intelligent:} $M' = \Omega(M, C, G)$ & 
    Meta-optimization through operator $\Omega$ that can redefine states, transitions, and goals based on context \\
    \bottomrule
  \end{tabular}
\end{table}
Each level of sophistication represents a step increase of dynamic capabilities, potentially accumulative. On top of the traditional DAG execution model (\textbf{Static}), the noisy and failure-prone real-world execution environment introduces the need for conditionals to recover and adjust course based on real-world observations (\textbf{Adaptive}). Though, this resulted in an explosion of if-then-else conditions that lead us to learning systems (\textbf{Learn}) where the environment can be learned without explicit programming. Then, we add goal-seeking behavior where the system also dynamically steers itself to an optimal state without explicit programming (\textbf{Optimizing}). Emerging AI capabilities represented by LLM- or LRM-based AI agents bring meta optimization capabilities where the state machine itself can be dynamically rewritten.

This intelligence hierarchy yields distinct operational trade-offs and its progression is driven by scientific requirements rather than technological possibility. Verification complexity increases from tractable for static $\delta$ to undecidable for metaoptimization~$\Omega$. Resource requirements scale from $O(1)$ lookups to potentially unbounded computation for intelligent reasoning. Learning systems require a data infrastructure to maintain history $H$, optimizing systems need an evaluation infrastructure for the cost function $J$, while intelligent systems demand sophisticated reasoning engines implementing $\Omega$ (e.g., LLM- and LRM-agents in Figure~\ref{figs:state_machine}). We can map this spectrum in existing systems: Traditional HPC workflows (\textbf{Static}), fault-tolerant frameworks with feedback $O$ (\textbf{Adaptive}), ML-guided parameter selection using learning $L$ (\textbf{Learning}), automated tuning platforms minimizing $J$ (\textbf{Optimizing}), and emerging autonomous lab controllers implementing the $\Omega$ (\textbf{Intelligent}).

\subsection{The Composition Dimension}
\label{subsec:composition_dim}

Composition defines how multiple state machines coordinate each other to achieve collective behavior. We identify five patterns with distinct properties (Table~\ref{tab:composition}) according to the progressive sophistication of coordination between entities, potentially in an accumulative fashion. Higher levels of sophistication in composition leverage dynamic coordination to navigate complex dependencies or unknown structures without manually encoding them.

\begin{table}[ht!]
  \caption{The composition dimension}
  \label{tab:composition}
  \footnotesize
  \begin{tabular}{p{3.8cm}p{4.0cm}}
    \toprule
    \textbf{Dimension} & \textbf{Description} \\
    \midrule
    \textbf{Single:} $M$ & 
    One isolated machine with no coordination \\
    \midrule
    \textbf{Pipeline:} $M_1 \circ M_2 \circ \ldots \circ M_n$ & 
    Sequential composition with unidirectional dataflow, enabling staged processing with clear dependencies \\
    \midrule
    \textbf{Hierarchical:} $M_{mgr.}(M_1, M_2, \ldots, M_n)$ & 
    Manager structure implementing delegation and supervision with centralized control \\
    \midrule
    \textbf{Mesh:} $\forall i,j: M_i \leftrightarrow M_j$ & 
    Full connectivity enabling peer-to-peer communication and collaborative problem-solving \\
    \midrule
    \textbf{Swarm:} $M = \Phi(\{m_1, m_2, \ldots, m_n\})$ & 
    Emergent behavior through emergence operator $\Phi$ transforming local interactions into global behavior \\
    \bottomrule
  \end{tabular}
\end{table}

Such coordination would start with a single standalone machine $M$ with no inter-workflow communication (\textbf{Single}) that can naturally progress to a sequential unidirectional data flow implemented with a sequential composition $M_1 \circ M_2$ where the output of $M_1$ feeds the input of $M_2$, enabling staged processing with clear dependencies (\textbf{Pipeline}). For more complex tasks, a manager pattern $M_{mgr}(M_1, …, M_n)$ where the manager orchestrates children supports divide-and-conquer strategies with centralized control (\textbf{Hierarchical}). Beyond, complex dependencies would require either a fully connected mesh $M_i \leftrightarrow M_j$ through message passing or a shared state to enable collaborative problem-solving (\textbf{Mesh}). Further, large problems with unknown structures would require emergent behavior via $\Phi({m_i})$ where simple local rules yield collective intelligence without central coordination~(\textbf{Swarm}).

Each pattern exhibits different scaling in terms of communication channels between entities. Pipeline composition $M_1 \circ M_2 \circ … \circ M_n$ requires $O(n)$ channels where hierarchical $M_{mgr}$ needs $O(n)$ channels per level. Mesh with $\forall i,j: M_i \leftrightarrow M_j$ demands $O(n^2)$ connections for all-to-all connections, but a swarm system using the emergence operator $\Phi$ would use only $O(k)$ local communications, where $k$ is the neighborhood size, to maintain scalability with a much larger number of participants. Existing implementations include batch processing (\textbf{Single}), multi-stage pipelines using the $\circ$ operator (\textbf{Pipeline}), workflow-of-workflows with $M_{mgr}$ (\textbf{Hierarchical}), collaborative platforms with full $\leftrightarrow$ connectivity (\textbf{Mesh}), and particle swarm optimization implementing $\Phi$ emergence (\textbf{Swarm}). Each level of sophistication profoundly impacts system design choices.

\subsection{The Evolution Matrix and Classification}

Following the intelligence and composition dimensions described in the previous sections, we create a $5 \times 5$ evolution matrix that provides us with a comprehensive taxonomy. Each cell represents a class example combining an intelligence level (static $\delta$, adaptive $\delta + F$, learning $L$, optimizing $\arg\min J$, intelligent $\Omega$) and a composition pattern (Single, Pipeline $\circ$, Hierarchical $M_{mgr}$, Mesh $\leftrightarrow$, Swarm $\Phi$) as depicted in Table~\ref{tab:5x5matrix}. This matrix helps in both defining a descriptive classification of systems or a prescriptive planning of trajectories. Current workflow systems cluster at the top-left with traditional DAGs at [Static $\times$ Pipeline] and fault-tolerant systems at [Adaptive $\times$ Pipeline]. The bottom-right frontier represents autonomous science. [Intelligent $\times$ Swarm] characterizes laboratories where meta-optimization $\Omega$ combines with emergence $\Phi$ for collective discovery.

\begin{table}[t]
  \caption{Representative Examples Across the 5×5 Evolution Matrix: Classification of Workflow Systems by Intelligence Level and Composition Pattern}
  \label{tab:5x5matrix}
  \footnotesize
  \begin{tabular}{rp{1cm}p{1cm}p{1cm}p{1cm}p{1cm}}
    \toprule
        & \textbf{Static} 
        & \textbf{Adaptive} 
        & \textbf{Learning} 
        & \textbf{Optimizing} 
        & \textbf{Intelligent}
        \\
        
    \midrule
    \textbf{Single} 
        & Script
        & Exception Handler
        & ML Model
        & Optimizer
        & LLM-Agent~\cite{significant2023autogpt, nakajima2023babyagi}
        \\
        
    \midrule
    \textbf{Pipeline} 
        & DAG
        & Conditional DAG
        & ML Pipeline
        & AutoML
        & Agent Chain~\cite{chase2022langchain, wu2023autogen}
        \\
        
    \midrule
    \textbf{Hierarchical}
        & Batch System
        & Dynamic Allocation
        & Ensemble
        & Hyper Optimization
        & Hierarchical Multi-Agent~\cite{rashid2018qmix}
        \\
        
    \midrule
    \textbf{Mesh} 
        & Fixed Grid
        & Load Balancing
        & Federated
        & Distributed Optimization
        & Agent Society~\cite{zhang2021marl}
        \\
        
    \midrule
    \textbf{Swarm} 
        & Parameter Sweep
        & Adaptive Sampling
        & Particle Swarm Opt.~\cite{kennedy1995pso}
        & Ant Colony~\cite{dorigo1996aco}
        & Emergent AI~\cite{warnat2021swarm}
        \\
        
    \bottomrule
  \end{tabular}
\end{table}

In this matrix, systems evolve by enhancing either intelligence or composition. Common trajectories include HPC workflows advancing from basic $\delta$ to feedback-enabled $\delta + O$ (Static $\rightarrow$ Adaptive), ML workflows progressing from isolated learning $L$ to pipelined $M_1 \circ M_2$ architectures with multiple models (Single $\rightarrow$ Pipeline), and Autonomous labs evolving from adaptive pipelines to optimizing hierarchies using $M_{mgr}$ with $\arg\min J$ capabilities. Not all systems need to target [Intelligent $\times$ Swarm] and should follow scientific needs. However, we foresee demands for higher degrees of autonomous behavior. Critical transitions towards such autonomy include adding learning $L$ (requires data infrastructure), implementing optimization $\arg\min J$ (needs objective specification), and achieving meta-optimization $\Omega$ (demands reasoning engines and knowledge bases).

The evolution towards autonomous scientific discovery requires coordinated advancement towards meta-optimization $\Omega$ and emergent coordination $\Phi$. The gap between current facilities with basic state transitions  $\delta$ and pipeline composition $\circ$ and [Intelligent $\times$ Swarm] can be interpreted as quite a revolution. However, the framework prescribes an evolutionary systematic progression in enhancing intelligence (e.g., Static $\delta$ $\rightarrow$ Adaptive $\delta+O$ $\rightarrow$ Learning $L$ $\rightarrow$ Optimizing $\arg\min J$ $\rightarrow$ Intelligent $\Omega$) within existing composition, then expanding coordination (Single $\rightarrow$ Pipeline $\circ$ $\rightarrow$ Hierarchical $M_{manager}$ $\rightarrow$ mesh $\leftrightarrow$ $\rightarrow$ Swarm $\Phi$). Each transition enables a new level of autonomy: Learning~$L$ brings adaptation, optimization $\arg\min J$ enables goal-seeking, meta-optimization $\Omega$ allows autonomous self-modification, while swarm emergence $\Phi$ achieves collective intelligence beyond individual capabilities.

\section{Challenges}
\label{sec:challenges}

In this Section, we explore the challenges in progressing in the journey along the dimensions of intelligence and composition. These challenges map to the AI community (intelligence) and the scientific workflows community (composition), and combined form a unique frontier in enabling AI to perform scientific discovery. A frontier of enabling autonomous scientific discovery.

\subsection{Challenges in AI for Scientific Discovery}
The physical-digital divide represents a fundamental barrier to autonomous science. AI in the physical realm is much more complex. Current generations of AI capabilities (e.g., LLMs, LRMs) excel at text correlation but lack causal understanding necessary for controlling irreversible experiments. This becomes problematic in a high-stakes environment with precious samples or expensive equipment. Validation, simulation, uncertainty quantification become critical. On top of this, heterogeneous vendor integration is challenging. Proprietary interfaces from different manufacturers prevent seamless automation. Remote orchestration across institutional boundaries adds communication latency and security concerns. Further, long-horizon autonomous operations face reliability challenges from a combination of error compounding, equipment failures, and environmental variations that AI must handle without human intervention.

Scientific AI requires multimodal understanding that goes beyond current text-based models. The gap between LLM capabilities and scientific data limits autonomous decision-making. Native understanding of simulations, sensor streams, and experimental observations is crucial. Diverse modalities demand new architectures that understand physical constraints, not just statistical correlations. On top of this, AI must reason about why experiments produce certain results, not merely recognize patterns. While pattern recognition is still useful in many cases, scientific AI must have the capability to operate at the frontier of human knowledge, balancing exploration and validity. Discoveries must be physically realizable, not just statistically probable. This challenge requires AI systems that comprehend both abstract theoretical concepts and concrete experimental constraints.

Zooming out to a global view of the distributed scientific complex with multiple capabilities (e.g., HPC, quantum, experimental facilities, observatories), multi-stakeholder alignment becomes a challenge in creating autonomous systems. Autonomous labs need to find an optimal balance between principal investigators prioritizing their own research, facilities maximizing throughput, and funding agencies demanding social impact. Without clear governance frameworks, AI systems may pursue efficiency over scientific merit or the other way around. Traditional concepts of intellectual property and credit assignment in publications will be challenged due to multi-agent collaborations and significant AI agent contributions. Resource allocation will be challenging with AI systems potentially learning how to game the system. New governance models enabled by relevant technologies must balance stakeholder interests while preserving scientific freedom to explore unexpected directions.

\subsection{Workflow Challenges}

\sloppy
Scientific workflows are increasingly composed of distributed elements that span instruments, edge devices, cloud platforms, and HPC systems across multiple facilities and institutions~\cite{antypas2021enabling, ferreiradasilva2024computer}. This transition from single-site pipelines to globally coordinated swarms introduces a new level of complexity in workflow composition~\cite{balaprakash2025swarm}. Each component may operate at different timescales, from microsecond-scale sensor measurements to month-long simulations and analyses. Managing coherence in such settings requires new abstractions for asynchronous coordination, fault tolerance, and adaptive resource utilization. Existing workflow systems, which often assume tightly coupled and statically defined execution environments, are ill-suited for managing emergent behaviors and loosely connected agents that collaborate across administrative boundaries.

\sloppy
As workflows adopt higher levels of intelligence, moving from adaptive systems to those capable of optimization and meta-reasoning, core assumptions about trust and reproducibility must be reconsidered. Traditional workflows enforce reproducibility through deterministic behavior and tightly controlled inputs. In contrast, intelligent workflows generate results through iterative adaptation, data-driven decisions, and evolving objectives. Reproducing such workflows may involve replicating the conditions and logic of decision processes rather than reproducing identical outputs. This shift calls for enhanced provenance models that can capture feedback mechanisms, learned behaviors, and context-sensitive decisions across agents~\cite{prov-agent}.
Provenance models need to evolve to support traceability of agent actions within the workflow context, enabling accountability, transparency, explainability, and auditability~\cite{prov_ech_continuum}.
Maintaining alignment with FAIR data principles~\cite{wilkinson2016fair, wilkinson2025applying} becomes more difficult when autonomous agents operate independently, select data dynamically, and pursue objectives that may change over time.

The emergence of agentic workflows in multi-institutional settings presents additional challenges related to interoperability, governance, and data compliance~\cite{autonomousscience}. Scientific campaigns increasingly require coordination across systems with differing policies, trust assumptions, and regulatory constraints. Agents acting on behalf of workflows must be capable of negotiating access, managing data across jurisdictions, and adhering to institutional governance models. Without common standards for capability description, data sharing, and execution intent, such workflows risk incompatibility and fragmentation. The governance of collective behavior becomes particularly challenging when workflows operate at the swarm level, where global outcomes emerge from the interaction of many locally autonomous agents. Future workflow infrastructure must embed mechanisms for policy enforcement, ethical guardrails, and transparent auditability to ensure that scientific freedom, compliance, and collaboration can coexist within intelligent and distributed ecosystems.

\subsection{Cultural, Adoption, and Ethical Challenges}

Cultural and adoption barriers pose the greatest threat in realizing autonomous science. There is a struggle in adopting transformation in the scientific community, as researchers trained in traditional methods resist trusting new AI-driven methods or ceding partial control to AI systems. In part, AI methods and AI systems have not earned the trust to fully be employed in scientific discovery, but this also fundamentally requires a mindset shift that challenges established practices and career incentives. With the industry years ahead in AI adoption, the scientific community risks obsolescence if cultural transformation fails. The community needs environments that foster AI collaboration while establishing evidence that autonomous systems can carry out discovery and augment scientists.

Ethical implications of AI-driven scientific discovery demand careful consideration of bias, access, and accountability. Bias in historical data may perpetuate existing biases about which research questions merit investigation. Equitable access to resources becomes critical to prevent creating “AI haves” versus “have-nots” in the scientific community. Further, when AI systems make costly errors by destroying samples or equipment, liability frameworks must clearly assign responsibility. Fundamentally, human agency and creativity in science must be preserved while leveraging AI capabilities, ensuring that automation enhances more than replaces human scientific insight and intuition. Without addressing these factors, technical advances alone cannot deliver the transformational potential of autonomous science.

\section{Architectural Evolution}
\label{sec:archvis}

To realize the evolutionary path outlined in the previous section, we introduce in this section an architectural vision for systems that support autonomous science. Rather than prescribing a single design, we propose a set of architectural patterns that accommodate varying levels of intelligence and composition, aligned with different stages along the evolution framework. The key principle is to enable a gradual and practical transformation of today's workflows into fully autonomous scientific systems in collaboration with AI autonomy. 
The following subsections describe guiding design principles, present a federated system architecture, and explore concrete application scenarios that illustrate how these concepts may be realized in practice.

\subsection{Design Principles}

A key requirement for architecting autonomous science systems is to support evolutionary transitions without disrupting existing capabilities. To that end, the architecture must accommodate a range of \textbf{levels of intelligence}, from static workflows to fully agentic systems, within a unified framework. This flexibility allows systems to operate with heterogeneous intelligence components while gradually adopting more sophisticated capabilities. \textbf{Backward compatibility} with existing WMSs is essential to preserve the significant investments and practices established over two decades of scientific workflow development. Furthermore, a \textbf{modular design} enables separation of concerns and the incremental adoption of autonomy-enabling components without requiring wholesale system rewrites. This evolutionary approach must also respect the foundational needs of scientific computing, including validation, reproducibility, provenance tracking, and auditability, ensuring trust and reliability throughout the transition.

To coordinate activities across diverse facilities while maintaining their operational independence, we propose a federated architecture. This design embraces \textbf{distributed control}, allowing each facility to retain autonomy over local policies, infrastructure, and instrumentation. Cross-facility coordination is enabled through \textbf{standard protocols} that support communication, capability advertisement, and resource discovery. These protocols facilitate dynamic matchmaking between agents, instruments, and services across administrative boundaries. Decoupling integration logic from local implementation details helps promote interoperability and scalability. This federated approach also aligns with practical constraints in national lab environments and global collaborations, where centralized orchestration is neither feasible nor desirable. The result is a flexible architecture that supports evolution toward autonomy while balancing integration needs with the realities of distributed scientific ecosystems.

\begin{figure}[t]
  \centering
  \includegraphics[width=\linewidth]{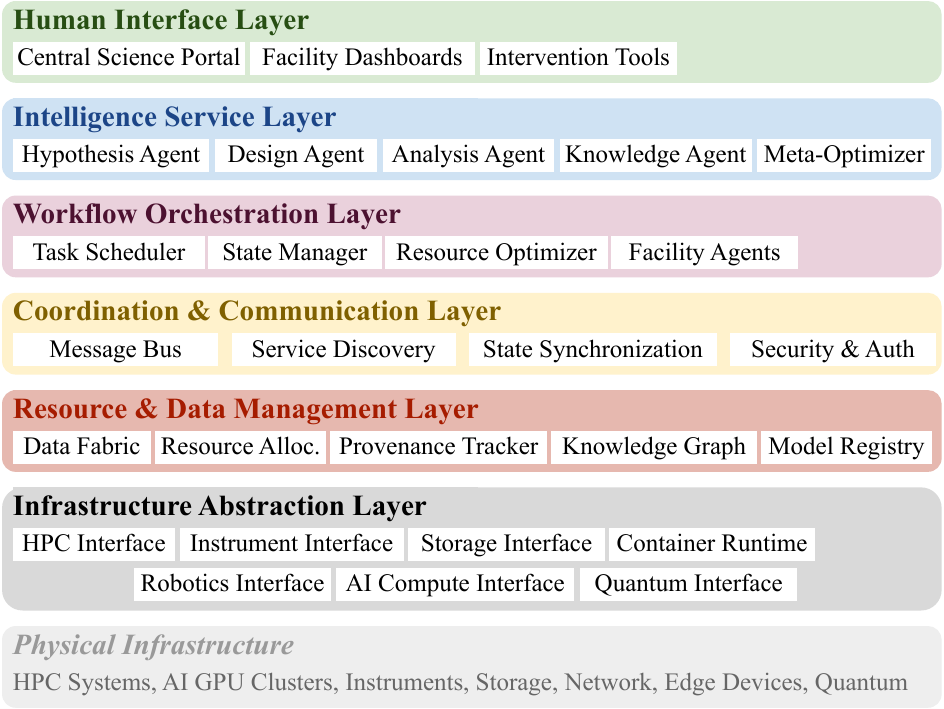}
  \caption{Architectural layers and components: Evolution introduces additional components in existing layers and additional layers for intelligence and coordination.}
  \label{figs:arch_layers}
\end{figure}

\subsection{Architecture}
\label{subsec:architecture}

Our architectural vision aims to materialize the evolution framework in Section~\ref{sec:framework} through systematic extension of the existing workflow infrastructure pursuing federated scientific discovery.  Figure~\ref{figs:arch_layers} illustrates how the current workflow architectures can gain new capabilities through additional components in existing layers and the introduction of an Intelligence Service Layer. We observe that each layer can evolve to support progressively sophisticated intelligence levels and composition patterns through focused extensions of additional considerations. Such enhanced layers orchestrate instruments, compute, data, and intelligence on top of a physical infrastructure with heterogeneous resources distributed across a scientific complex.

\subsubsection*{Human Interface Layer:}
Traditional static dashboards evolve into dynamic planning and intervention tools supporting human-AI collaboration implementing human-in-the-loop (e.g., human intervention in every step) or human-on-the-loop (e.g., minimal human intervention) interaction in autonomous scientific discovery. Beyond monitoring, interfaces become a key component in enabling near real-time steering of autonomous experiments, validation of AI-generated hypotheses, and intervention when agents approach decision boundaries. New categories of user interface tools such as an integrated development environment (IDE) for human-AI scientific collaboration will emerge specifically designed for planning, experiment designing, knowledge browsing, and intervention.

\subsubsection*{Intelligence Service Layer:}
This new layer houses capabilities for autonomous agents that implement the intelligence dimension. Potentially leveraging service-oriented architectures or microservices, agents will be abstracted as long-running tasks or micro-workflows asynchronously triggered by standard API interactions enabling composition across intelligence levels and would leverage and orchestrate the resources with the capabilities exposed by the other layers. Hypothesis agents generate novel research directions. Design agents propose experimental configurations. Analysis agents interpret results. Knowledge agents (e.g., scientific knowledge graphs) maintain contextual understanding. Meta-optimizers dynamically reconfigure entire workflows.

\subsubsection*{Workflow Orchestration Layer:}
Traditional schedulers are extended with facility agents (potentially served from the intelligence layer) that consider local capabilities and constraints when interrogated. State managers would track learning progress and hypothesis evolution across armies of agents. Resource optimizers need to balance immediate experiment needs with long-term discovery goals. Ultimately, the layer must support conditional execution based on near real-time results, dynamic workflow modification through meta-optimization feedback, and coordination mesh and swarm behaviors.

\subsubsection*{Coordination \& Communication Layer:}
Message buses will evolve to support semantic agent negotiation on top of protocols like AMQP 1.0~\cite{standard2012oasis} for federated event-driven workflows. Service discovery extends standards like OGSA (Open Grid Services Architecture~\cite{talia2002open}) for cross-facility interoperability. WSRF (Web Services Resource Framework~\cite{foster2005modeling}) enables stateful interactions that can manage distributed learning states and progress. Security frameworks like Globus Auth~\cite{tuecke2016globus} can be extended to authenticate inter-agent communication. Scalable consensus protocols for multi-agent decision-making and distributed state management are required and should provide audit trails for autonomous actions across federated infrastructures.

\subsubsection*{Resource \& Data Management Layer:}
Data fabrics leverage data transfer services like Globus Transfer~\cite{allen2012software} for high-performance movement of multimodal scientific data across facilities. Knowledge graphs represent relationships between hypotheses, experiments, and results, synchronized across sites with eventual consistency. 
Provenance tracking will extend to capture AI reasoning chains and swarm emergence patterns while creating the relationships with these other concepts in the knowledge graph.
Model registries version both AI/ML models and various AI input artifacts such as experimental protocols. Resource allocation implements dynamic service-level agreements for cross-facility negotiation, considering compute availability, sample scarcity, and exploration-exploitation trade-offs.

\subsubsection*{Infrastructure Abstraction Layer:}
Heterogeneous resources will be abstracted through unified interfaces via standards like the grid computing standards~\cite{talia2002open}. New abstractions should support AI-specific hardware (TPUs, low-precision GPUs), robotic systems, and quantum devices with both interactive and batch usage models. Container runtimes must adapt for long-running AI services with model persistence. Specialized interfaces are required to manage real-time instrument control, streaming data, asynchronous experiment monitoring, and hybrid classical-quantum workflows.

\begin{figure}[t]
  \centering
  \includegraphics[width=\linewidth]{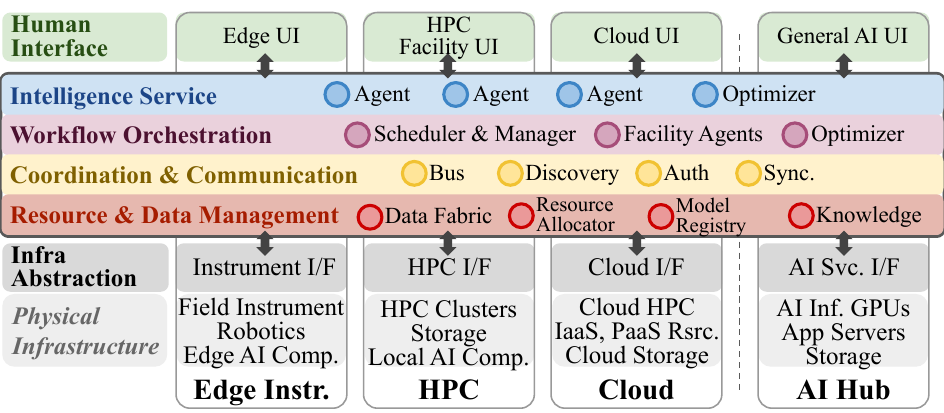}
  \caption{Deployment of architectural components in a federated environment: The modern distributed scientific environment is extended with additional layers and an AI hub facilities for AI inference specialized compute and storage.}
  \label{figs:arch_dep}
\end{figure}

\begin{figure*}[t]
  \centering
  \includegraphics[width=16cm]{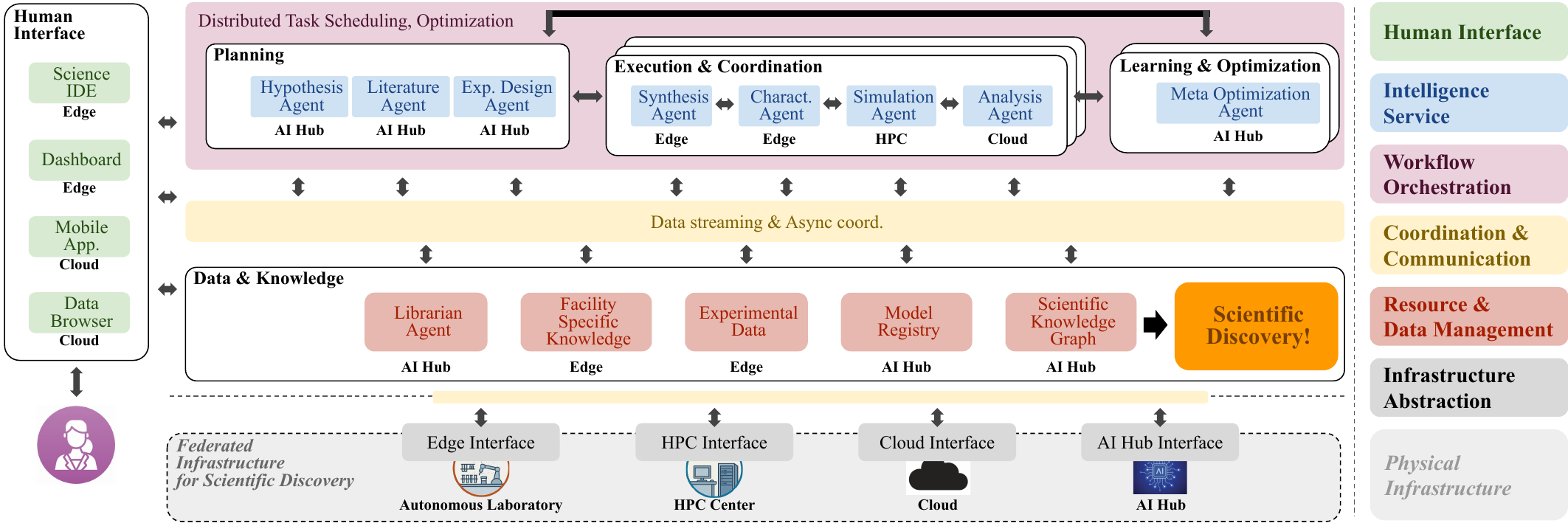}
  \caption{Example of federated scientific discovery: components in the intelligence layer distributed across the infrastructure iterates through various hypothesis in real time.}
  \Description{Description of the image}
  \label{figs:example-flow}
\end{figure*}

\subsection{Deployment}
\label{subsec:deployment}

\subsubsection*{Federation in the scientific environment:}
Figure~\ref{figs:arch_dep} illustrates how the layered architecture is deployed across facilities in federation. Architectural components are distributed across and are coordinated while each facility maintains operational autonomy with components based on local specialization. Materials synthesis labs emphasize robotic interfaces, while HPC centers focus on simulation and optimization services. Standard protocols enable advertisement and dynamic service discovery across boundaries. Each facility operates at different evolution levels with each component abstracted as an execution unit but with different capabilities seamlessly interacting with each other. Deployment patterns of intelligence will range from edge devices providing sub-second inference at instruments to regional AI hubs coordinating thousands of parallel campaigns, providing balance between latency, cost, and capability.

\subsubsection*{Physical Infrastructure - The AI Hub Extension:}
AI hubs represent a critical new infrastructure distinct from traditional HPC systems. While HPC emphasizes double-precision floating-point for physical simulations, AI inference requires high-throughput, lower-precision arithmetic (i.e., FP16/INT8) with massive memory bandwidth, especially with the advent of transformer-based models. AI training and large-scale swarm intelligence coordinating thousands of agents demand high-speed interconnects (>400Gbps interconnect) for real-time consensus within inference clusters, while hundreds of agents must operate efficiently across distributed facilities (>100Gbps networks). The required investment scales from single-rack edge deployments for local inference to multi-megawatt AI supercomputers for AI inference scaling and swarm coordination.

\subsection{Federated Autonomous Scientific Discovery}
\label{subsec:example}

Consider materials discovery progressing through the federated infrastructure in Figure~\ref{figs:example-flow}. Scientists interact via the Science IDE to initiate campaigns and interact with the agents. With specifications about the experimental process, planning agents at the AI Hub generate hypotheses, review literature, and design experiments. These plans will submit tasks to multiple instances of execution agents distributed across facilities coordinating tasks such as synthesis at edge laboratories, characterization at beamlines, simulation on HPC clusters, and analysis in the cloud. Results from each agent will stream through the data fabric and trickle into the knowledge graph where the meta-optimization agent refines strategies. Such a continuous loop will operate autonomously across institutional boundaries with agents coordinating through asynchronous messaging and state synchronization methods while humans monitor progress via dashboards or mobile devices, intervening only when needed. In its ultimate autonomous form, all with no manually defined DAGs in place.

\subsection{Roadmap}
\label{subsec:roadmap}

Today’s scientific workflows primarily operate at the [Static $\times$ Pipeline] with emerging [Adaptive $\times$ Pipeline] capabilities. The intelligence gap spans from current adaptive $\delta+O$ to meta-optimization $\Omega$ necessary, while composition must evolve through pipelines to swarm coordination with emergence operator $\Phi$. Projects like FireWorks~\cite{jain2015}, Pegasus~\cite{deelman2019}, and Parsl~\cite{babuji2019} provide robust pipeline orchestration with limited adaptive features through conditional execution. Some systems achieve basic learning through ML integration or optimization via parameter sweeps. Hierarchical composition exists in meta-workflows, but true mesh or swarm coordination remains experimental. Data fabric building blocks such as Globus~\cite{allen2012software, tuecke2016globus} and similar services enable data movement, but lack coordination capabilities for mesh and swarm behaviors.

Achieving the federated autonomous discovery requires systematic infrastructure extensions built on top of existing infrastructure, charting a trajectory of evolution from today’s limited automation to tomorrow’s autonomous science. At the coordination \& communication layer, standards must evolve to support stateful AI services and semantic agent communication. Authentication and transfer services need augmentation with capability negotiation protocols assuming non-human access scenarios. Workflow engines require integration with reasoning systems for meta-optimization ($\Omega$) implementation for dynamic goal and \textit{state machine} modification. Scalable and standard consensus algorithms must enable swarm coordination ($\Phi$) across hundreds or thousands of agents. For resources \& data management, critical additions include persistent model registries for learning systems $L$, objective specification frameworks and validation frameworks for $\arg\min J$, knowledge graphs linking experiments to theories.

\section{Opportunities}
\label{sec:opportunities}

Transformative opportunities for scientific discovery emerge as workflows progress towards higher intelligence and composition. By systematically advancing along both dimensions, the scientific community can unlock capabilities that reshape how we understand and interact with the natural world.

\subsection{Autonomous Discovery}
\label{subsec:autonomous_discovery}

The shift to autonomous discovery presents a pivotal opportunity to redefine the nature and pace of scientific research~\cite{pauloski2025empowering, autonomousscience}. Self-driving laboratories operating 
with minimum human oversight
are no longer aspirational. They are already enabling new modes of exploration~\cite{abolhasani2023rise, hysmith2024future}. Autonomous materials discovery campaigns have evaluated over one million candidate compounds, demonstrating the ability to scale scientific experimentation far beyond traditional approaches. Workflows that once required extensive manual coordination across multiple facilities are now transitioning into intelligent systems capable of orchestrating activities across institutional and geographic boundaries. This transformation reflects a broader rethinking of workflow models, where execution engines evolve into intelligent agents that reason, adapt, and act based on scientific context. These capabilities allow researchers to move beyond rigid protocols and instead focus on strategic decision-making, unlocking the potential to address challenges in climate, health, and energy with greater speed and flexibility.

Autonomous discovery also marks a departure from traditional workflow automation, enabling the formation of complete discovery loops~\cite{lordan2024taming}. Instead of executing fixed computational tasks, these systems integrate robotic synthesis, near real-time characterization, and modeling within iterative and adaptive cycles. Through meta-optimization, workflows gain the ability to redefine their goals in response to new data and evolving scientific questions. Rather than simply reacting to inputs, they initiate new hypotheses and determine how best to test them. When deployed across multiple facilities, autonomous agents coordinate their actions to explore diverse scientific questions in parallel and dynamically exchange insights~\cite{balaprakash2025swarm}. This model of distributed and intelligent discovery establishes a foundation for science that is more responsive, collaborative, and scalable, creating a path toward continuous innovation that is both efficient and inclusive.

\subsection{Accelerated Time to Discovery}
\label{subsec:accelerated_discovery}

With higher degrees of autonomy, we foresee opportunities in
significantly reducing
human bottlenecks in the experimental cycle, bringing the potential of 100-fold acceleration in scientific discovery~\cite{autonomousscience}. Current discovery pipelines stall at points waiting for researchers to analyze data, design next experiments, or coordinate resources that are distributed across facilities. Learning systems ($L$) reduce iteration time by recognizing patterns in experimental outcomes, and optimizing ($\arg\min J$) automatically tune parameters to maximize discovery likelihood. For example, in drug discovery, traditional pipelines requiring years of manual iteration could be compressed to weeks when AI agents continuously analyze results, adjust molecular structures, queue synthesis reactions, and perform experiments with robots without human intervention.

In a distributed modern scientific environment, scientific campaigns run across multiple experimental facilities and compute facilities. Intelligence at the edge transforms how we deliver computational power to scientists, representing a shift from centralized HPC to distributed intelligence embedded throughout the research infrastructure. Mesh and swarm composition enable parallel exploration of complex problem spaces, with each node contributing specialized expertise including experimental facilities and observatories interacting with the real world. When combined with faster simulation code generation, experiment design and data analysis significantly compress the entire discovery cycle with humans in the loop. This distributed intelligence model democratizes access to AI capabilities, ensuring every researcher can leverage autonomous systems regardless of their computational resources.

\subsection{New Scientific Methods and Solutions}
\label{subsec:new_science_methods_solutions}

Autonomous scientific discovery will bring a paradigm shift from hypothesis-testing science to hypothesis-generating science that expands the frontier of discoverable knowledge. AI-driven hypothesis generation has the potential to transcend human limitation in exploring non-obvious connections. While traditional science is limited by human intuition, intelligent systems with multimodal capabilities can discover patterns potentially invisible to humans. In material science, AI agents can identify counterintuitive combinations of elements that violate conventional wisdom yet exhibit remarkable properties.

Swarm intelligence enables emergent scientific insights through collective discovery at a larger scale. A large population of AI agents can simultaneously explore different areas of complex problems at scale, leveraging the emergent phenomena. With agent collaboration defined by the swarm operator, the emergent collective behavior ($\Phi$) coordinating meta-optimization  ($\Omega$) creates scientific exploration that no single agent could achieve. In drug discovery or chemistry, large-scale swarm intelligence explores vast solution spaces uncovering promising combinations at accelerated speed. By systematically exploring regions of parameter spaces that humans would never consider, AI systems can unveil entirely new classes of materials, reactions, and phenomena.

\subsection{Science as Leadership in AI}
\label{subsec:differentiation}

The scientific community is uniquely positioned to lead the autonomous science (r)evolution through specialized AI development. Unlike the industry focusing on business and enterprise-focused automation and optimization problems, scientific AI must handle causality, physical constraints, and uncertainty quantification.
These requirements position national laboratories, universities, and research institutes as innovation leaders in AI. Integrated facilities, scientific domain expertise, and capital for fundamental research create an ecosystem unmatched by commercial-only entities. The Autonomous Interconnected Science Lab Ecosystem (AISLE) grassroots network~\cite{autonomousscience} represents this leadership by connecting autonomous capabilities across institutions to establish global standards for reproducible, validated AI-driven science. Such movements in autonomous scientific discovery help the community to shape how AI transforms research practice. By developing specialized AI systems that understand scientific principles, we ensure AI capabilities fully augment scientists.

Building the scientific AI ecosystem requires strategic partnerships that leverage industry capabilities while maintaining scientific integrity. Instead of competing against large capital investments of the AI industry, the scientific community positions itself as a validation partner and co-developer for industry AI solutions. Domain-specific foundation models trained and validated on scientific data and methods provide capabilities the industry would not develop, while creating testbeds for safe AI experimentation in physical sciences allows rapid iteration while preventing costly errors. Such partnerships address the current challenges in AI and scientific discovery, forming a unique frontier that can take the capabilities of industry AI solutions to the next level as well as enabling scientific breakthroughs through accelerated science.

\section{Strategic Bets}

\subsubsection*{AI Research}
Investment in AI does not mean merely adopting industry solutions. The scientific community must invest strategically in AI capabilities purpose-built for discovery. Critical developments include scientific foundation models that understand causality and physical constraints, swarm intelligence frameworks enabling distributed discovery, and multimodal AI that natively processes simulations, experimental data, and theoretical models. Community-driven standards through grassroots networks like AISLE ensure interoperability and would take us close to the vision of intelligence distribution, providing AI assistance to every scientist. Such advancements require partnerships with the industry while maintaining scientific rigor co-exploring the unique frontier in AI and science. AI research priorities must focus on handling research uncertainty, reasoning about physical laws, and generating testable hypotheses beyond human intuitions. This research must be based on safe experimentation frameworks for AI in physical sciences and new validation methods for AI-discovered knowledge, enabling human-on-the-loop autonomous systems. Success in this area demands a balance between the rapid adoption of industry advances and the development of science-specific capabilities that industry will not prioritize.

\subsubsection*{Workflows Research}
To realize the full potential of agentic workflows, the workflows research community must commit to foundational shifts that move beyond traditional orchestration models. Future systems should support the complete intelligence and composition spectrum, encompassing static, adaptive, learning, optimizing, and intelligent workflows organized across increasingly collaborative execution patterns. Workflow management systems will need to evolve into modular, intelligence-aware platforms that integrate components operating at different cognitive levels while preserving compatibility with existing DAG-based tools. FAIR-compliant data infrastructure, distributed state management, and advanced provenance systems are essential to support the coordination, adaptability, and traceability of learning-enabled workflows. Communication protocols between agents must be standardized to enable transitions from pipeline-based systems to fully emergent swarms. These protocols should be demonstrated through reference implementations that show how workflows can incrementally evolve through the intelligence and composition matrix. Vendor-agnostic interfaces, hybrid execution support, and community-driven specifications will help ensure these systems can operate across institutional boundaries without sacrificing existing investments.

\subsubsection*{Infrastructure and workforce investments}
Investment in infrastructure and workforce development is equally critical to support this evolution. Infrastructure must include high-bandwidth networking for near real-time coordination across facilities, specialized AI accelerators located near experimental instruments, and edge computing platforms capable of executing intelligent agents close to data sources. Software platforms are needed to enable secure, collaborative AI workflows, simulation environments for validating agent behaviors prior to deployment, and standardized APIs that connect AI reasoning with physical instruments and computational environments. Shared testbeds such as those promoted by the AISLE initiative will allow communities to validate autonomous systems in controlled, reproducible settings. These testbeds serve as essential enablers of cross-institutional collaboration and reduce barriers to adopting automation. A complementary focus on workforce development is needed to create training programs that blend domain expertise with AI fluency, establish new career paths for AI-science practitioners, and foster communities capable of co-evolving with autonomous systems. These investments ensure that the scientific ecosystem is prepared not only to adopt autonomy, but to shape its future.

\section{Discussion and Conclusion}
\label{sec:conclusion}

In this paper, we presented a conceptual framework that helps us unify traditional scientific workflows and emerging AI agents through their common foundation, the state machine as their execution model. This enabled us to introduce two evolutionary dimensions which are intelligence (from static to intelligent) and composition (from single to swarm) creating a $5 \times 5$ matrix. This matrix serves as both a taxonomy for existing systems and a roadmap towards autonomous science. With this framework, we reveal that the path to autonomous discovery is evolutionary rather than revolutionary, enabling systematic progression from current DAG-based workflows to meta-optimizing and swarm-coordinated systems. Our architectural blueprint shows how we can build upon the federated foundations to support this transformation in an evolutionary way. This evolution promises a clearer path to accelerate scientific discovery by factors of 10-100x through intelligent coordination across distributed facilities.

Future work must focus on building the foundational infrastructure, protocols, and abstractions that enable workflows to operate seamlessly across the intelligence-composition spectrum and heterogeneous facility landscapes. Advancing this vision will require co-design of agentic workflow systems alongside AI-driven planning, semantic metadata models, and learning-enabled optimization components that can reason about scientific goals, resources, and uncertainty. Establishing robust testbeds for validating progressive levels of autonomy, as well as defining benchmarks and reference implementations, will be essential for tracking system evolution and ensuring reproducibility. These efforts must also include the development of formal representations for scientific intent, unified interfaces for facility integration, and governance frameworks for safe and collaborative deployment of autonomous agents. Ultimately, realizing this ecosystem will require sustained collaboration across domain scientists, computer scientists, and infrastructure providers to ensure that future workflows are not only more intelligent and adaptable, but also broadly usable, trustworthy, and aligned with evolving scientific practice.

\section*{Acknowledgment}
This research used resources of the OLCF at ORNL, which is supported by DOE's Office of Science under Contract No. DE-AC05-00OR22725.


\end{document}